\documentclass{svmult} 

\usepackage{amsmath}
\usepackage{amsfonts}
\usepackage{latexsym}
\usepackage{graphicx}
\usepackage{url}

\newcommand{\mynorm}[1]{\left|\left|#1\right|\right|}

\begin{document}

\title*{SparseCodePicking: feature extraction in mass spectrometry 
using sparse coding algorithms}
\titlerunning{SparseCodePicking: feature extraction in mass spectrometry}

\author{Theodore Alexandrov\inst{1}, 
        Klaus Steinhorst\inst{1},
        Oliver Kesz\"ocze\inst{1},
        Stefan Schiffler\inst{1}}
        
\authorrunning{Alexandrov et al.}

\institute{Center for Industrial Mathematics, University of Bremen, 
Bibliothekstr. 1, D-28334 Bremen, Germany, 
\texttt{theodore@math.uni-bremen.de}}

\maketitle

\begin{abstract}
Mass spectrometry (MS) is an important technique for chemical 
profiling  which calculates for a sample a high dimensional histogram-like spectrum. 
A crucial step of MS data 
processing is the peak picking which selects peaks containing 
information about molecules with high concentrations which are of 
interest in an MS investigation. We present a new procedure of 
the peak picking based on a sparse coding algorithm.
Given a set of spectra of different classes, 
i.e. with different positions and heights of the peaks,
this procedure can extract peaks by means of unsupervised learning.
Instead of an $l_1$-regularization penalty term used in the original
sparse coding algorithm we propose using an elastic-net 
penalty term for better regularization. The evaluation is done by means of simulation. 
We show that for a large region of parameters the proposed peak picking 
method based on the sparse coding features outperforms a mean spectrum-based method.
Moreover, we demonstrate the procedure applying it to two real-life datasets.

\end{abstract}

\section{Introduction}
\label{sec:AKSS09_intro}
Mass spectrometry (MS) is an important technique for chemical profiling 
and is a major tool in proteomics, a discipline interested in large-scale 
studies of proteins  expressed by an organism. In medicine, proteomics (and MS respectively)
contributes to clinical research by identification of biomarker proteins related to a disease,
e.g. produced by a tumor tissue or by the immune system in response to a disease. 

Given a sample of e.g. serum or urine, an MS instrument produces a high dimensional 
histogram-like spectrum. The high sharp peaks of this spectrum contain information 
about moleculae with high concentrations. By means of a peak picking procedure a raw spectrum is 
transformed to the so-called line spectrum which is a list of positions and intensities of 
the detected peaks. Although there exist MS data processing approaches working directly with the 
raw spectra, it is more usual in proteomics to work with line spectra 
since in an ideal case they provide all the necessary information about a spectrum in a significantly 
compressed form. 

Therefore, peak picking is usually a starting and a crucial step of MS data analysis. 
There exist various methods of peak picking but they are still far from perfect~\cite{AKSS09_YangHY09}. 
Nevertheless, the ubiquity of the peak picking problem as well as new MS technologies and
challenges still motivates many researchers to develop new methods for its solution.

This paper presents a new procedure of peak picking based on the sparse 
coding algorithm of~\cite{AKSS09_Lee06}. 
This procedure is motivated by recent successes of algorithms based on a sparse
representation of the data, also in MS, see e.g.~\cite{AKSS09_bredies09}. 

The key idea of the sparse coding is to represent vectors of a matrix
in a linear subspace where not only the coefficients but also the basis vectors of the subspace
are optimized. The optimization is performed simultaneously minimizing (1) the distance between 
the original data and the found approximation and (2) $l_1$-norm of the coefficients used 
in the representation. As usual in sparsity-based methods, $l_1$-norm is used as an approximation
of the $l_0$-norm (the number of non-zero coefficients) because optimization of the latter 
is an $NP$-hard problem.
. Since an optimization problem
involving $l_0$-norm is \emph{NP}-hard, an $l_1$-penalty is used for which optimization algorithms exist.

Let us have a dataset of spectra with common peaks. We suppose that the dataset represents
several classes, where one class is characterized by a set of peaks with the same positions and
similar heights. Note that for applying our procedure, we do not require
class-labels of spectra or the number of classes. 

Given a set of spectra, our procedure of peak picking is as follows. 
First, the sparse coding algorithm finds a sparse approximation of the dataset. Each spectrum
is represented using a few basis vectors (of the same dimension as the spectra) 
which capture the common features of the spectra. Then a simple peak picking procedure 
is applied on each basis vector found. Finally, we do an additional check reducing 
the number of false positives.

In Section~\ref{sec:AKSS09_method} we present our peak picking procedure using an improved sparse
coding algorithm with an elastic net penalty term. In Section~\ref{sec:AKSS09_evaluation}
we evaluate the procedure applying it to simulated datasets and providing mean accuracy 
characteristics. The procedure is compared with a simple mean spectrum-based peak picking procedure.
Moreover, we provide results of processing two real-life datasets.
Section~\ref{sec:AKSS09_concl} concludes the paper.

\section{Proposed procedure of peak picking}
\label{sec:AKSS09_method}

\subsection{Feature extraction using sparse coding}
Let us have a dataset of $R$ spectra of length $L$ which belong to $D$ classes
($D\ll R$)
each characterized by common peaks at the same positions and with similar 
heights. Given the matrix $\mathbf{X}\in\mathbb{R}^{L \times R}$ with spectra in columns,
a sparse coding algorithm represents each spectrum (a column of $\mathbf{X}$) in terms 
of some basis where the coefficients are optimized to produce a sparse representation.
In the sparse coding algorithm of~\cite{AKSS09_Lee06}, this is achived by solving
the following optimization problem
\begin{eqnarray}
\label{eq:AKSS09_sparsecoding}
   \min_{B,S} &\frac{1}{2}\mynorm{\mathbf{X-BS}}^2_\text{F} + \alpha\sum_j\mynorm{S_j}_1,\\
&\text{subject to} \mynorm{B_{j}}_2^2 \leq C,~~~~~~~~
\end{eqnarray}
with respect to a matrix of basis vectors $\mathbf{B}\in\mathbb{R}^{L \times L}$ and 
a matrix of the corresponding coefficients $\mathbf{S}\in\mathbb{R}^{L \times R}$,
where $\mynorm{\cdot}_\text{F}$ is the matrix Frobenius norm, $\mynorm{\cdot}_1$ is the vector $l_1$-norm, 
and $\mynorm{\cdot}_2$ is the vector euclidean norm; $S_j$ and $B_j$ denote the $j$-th column
of $\mathbf{S}$ and $\mathbf{B}$, respectively.
The hyperparameters of the optimization problem are the regularization parameter $\alpha$ 
and the boundary on the norm of basis vectors $C$. 

The minimization problem (\ref{eq:AKSS09_sparsecoding}) is solved in two steps. First,
we learn the coefficients $\mathbf{S}$ keeping the basis fixed using the Feature Sign Search (FSS) 
algorithm minimizing~(\ref{eq:AKSS09_sparsecoding}) for a fixed $\mathbf{B}$, 
then for the learned coefficients we optimize the basis $\mathbf{B}$ using the Lagrange dual.
For more details see~\cite{AKSS09_Lee06}.

Finally, for each column $X_j$ of $\mathbf{X}$ we have a sparse representation in basis $\mathbf{B}$
containing only a few basis vectors $B_j$ ($j\in\mathcal{I}$), where $\mathcal{I}$ is a set of
indices of non-zero rows of the matrix $\mathbf{S}$.

\subsubsection{Proposed elastic-net regularization}

Applying the described algorithm to a simulated dataset (see Section~\ref{sec:AKSS09_evaluation})
we found out that in many cases the regularization used in (\ref{eq:AKSS09_sparsecoding}) 
is not enough to provide any reasonable results. To cope this problem we introduced an additional
$l_2$-penalty term that leads to the so-called elastic-net regularization with the following optimization problem
\begin{eqnarray}
\label{eq:AKSS09_elasticsparsecoding}
   \min_{B,S} &\frac{1}{2}\mynorm{\mathbf{X-BS}}^2_\text{F} + \alpha\sum_j\mynorm{S_j}_1 + \beta \sum_j\mynorm{S_j}_2^2,\\
&\text{subject to} \mynorm{B_{j}}^2 \leq C.~~~~~~~~~
\end{eqnarray}
For solving this problem, we improved the FSS algorithm, see~\cite{AKSS09_schifflerSPARS09} for more details.

\subsection{Peak detection in the features extracted}
\label{sec:AKSS09_peakdetection}
In general, the basis $\{ B_j \}_{j\in\mathcal{I}}$ represents a compressed version of $\mathbf{X}$.
In our case, since the peaks have large values compared to other regions of spectra and 
they are common in each class of spectra, it is natural to expect them to be 
presented in $\{ B_j \}_{j\in\mathcal{I}}$ and this is what we observe in our applications 
(see Section~\ref{sec:AKSS09_evaluation}). Moreover, only the common features should be presented 
in the basis thus providing (1) denoising and (2) removal of spurious (not class-relevant) 
peaks corresponding to chemical compounds which do not express difference between the classes.

Taking this into account, we propose to do peak picking based not on raw 
spectra as it is normally performed, but on basis vectors $\{ B_j \}_{j\in\mathcal{I}}$. 

In this paper, we applied to $\{ B_j \}_{j\in\mathcal{I}}$ a simple peak picking algorithm using 
the Matlab function \emph{findpeaks} (Matlab R2008a, Signal Processing Toolbox).
First, we normalized each vector $B_j$. Then for a vector the peak positions were found 
as local maxima whose height is 2.5 larger than the mean value of the vector. 
Other values of multiplier (2 and 3) as well as the median value used instead of the mean value
have been tested producing almost identical results.

\subsubsection{Reduction of false positives} 

One could expect that, for a small number of classes, the sparse coding algorithm provides 
only one basis vector $B_j$ for each class ($\mathcal{I}$ of size $D$). 
However, examining results of processing the simulated data (see Section~\ref{sec:AKSS09_evaluation}), 
we realized that for many settings more vectors $B_j$ are found. 
We remarked that in this case some redundant vectors contain structural information presented 
in one or more classes, but some of them are very noisy. Definitely, any peak picking applied
on such vectors increases the number of false positives, i.e. the spikes identified as peaks 
but not belonging to the ground-truth peaks. We suppose that such noisy basis
vectors are produced because of insufficiently strong $l_1$-penalization. Another reason 
could be the sparse coding algorithm itself because though at each step (optimization of 
$\mathbf{S}$ and $\mathbf{B}$) a convex problem having an unique solution is solved, 
the problems \ref{eq:AKSS09_sparsecoding} and \ref{eq:AKSS09_elasticsparsecoding} do not have global solution.

To reduce the number of false positives, we examined each peak found separately, checking whether 
the area under peak is large enough. For this check we pre-specify the minimal possible width of a peak. 
When processing a real-life dataset, we estimated the minimal possible width of a peak after
visual examination of several large peaks.

\begin{figure}%
\centerline{\includegraphics[width=0.99\textwidth]{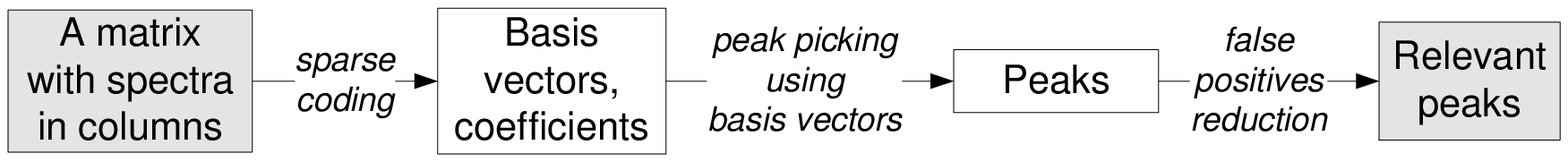}}%
\caption{The general scheme of the proposed peak picking procedure.}%
\label{fig:AKSS09_genscheme}%
\end{figure}

The general scheme of the proposed procedure is presented in Fig.~\ref{fig:AKSS09_genscheme}.

\section{Evaluation}
\label{sec:AKSS09_evaluation}
\subsection{Simulated data}
We evaluated the proposed peak picking procedure using simulated data. 
Each spectrum has been simulated using the following additive model. 
First, several Gaussian peaks were simulated on pre-specified positions with slightly
variable heights. Second, spurious peaks of smaller heights were added at random 
positions. Third, a white Gaussian noise was added throughout a spectrum.

\begin{figure}%
\hspace{0pt}
\centerline{
\begin{minipage}{7pt}
\large \textbf{a} \vspace{80pt}
\end{minipage} 
\begin{minipage}{0.7\textwidth}
\includegraphics[width=\textwidth]{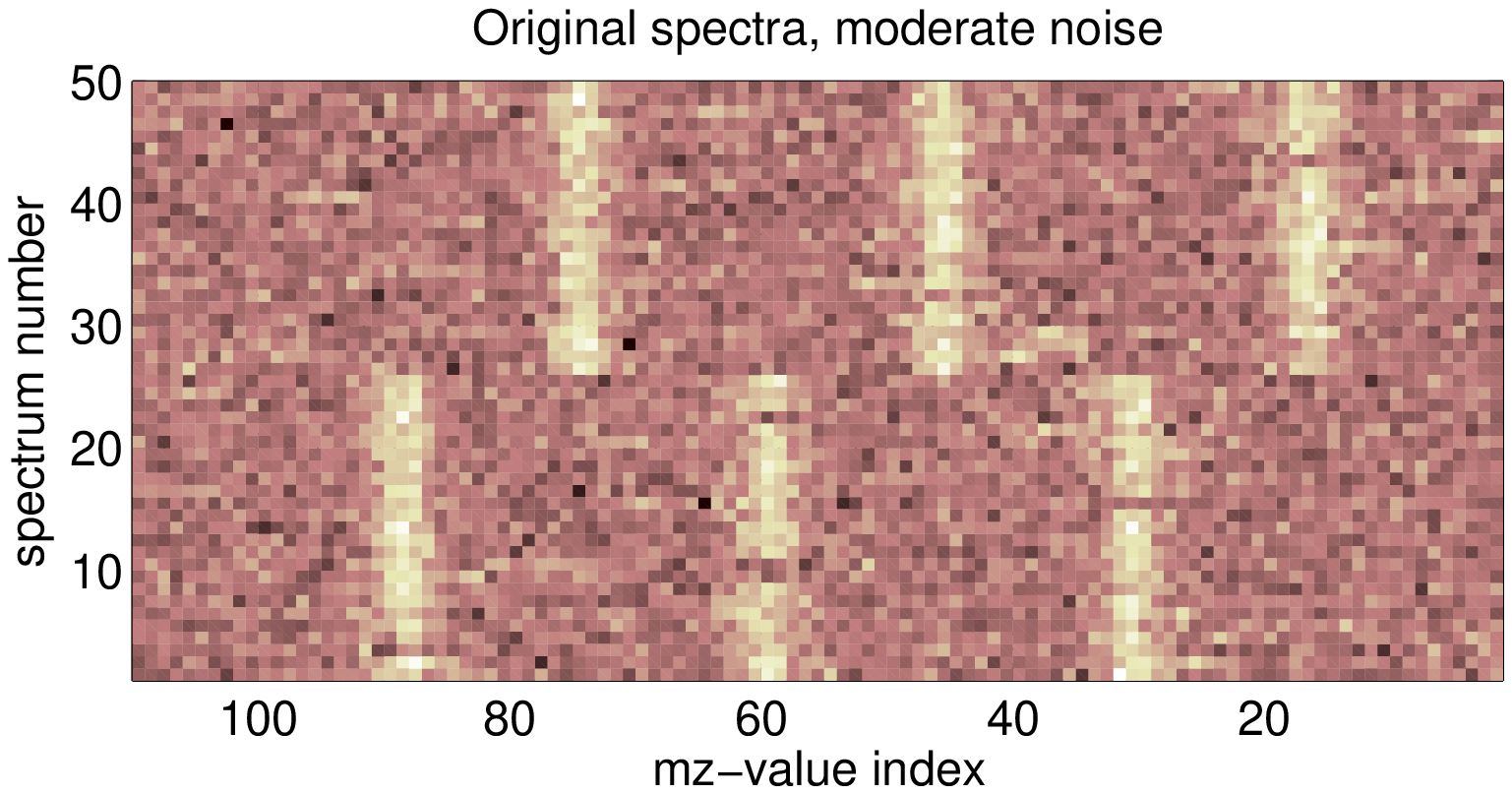}
\end{minipage}}
\hspace{0pt}

\centerline{
\begin{minipage}{7pt}
\large \textbf{b} \vspace{20pt}
\end{minipage} 
\begin{minipage}{0.7\textwidth}
\includegraphics[width=\textwidth]{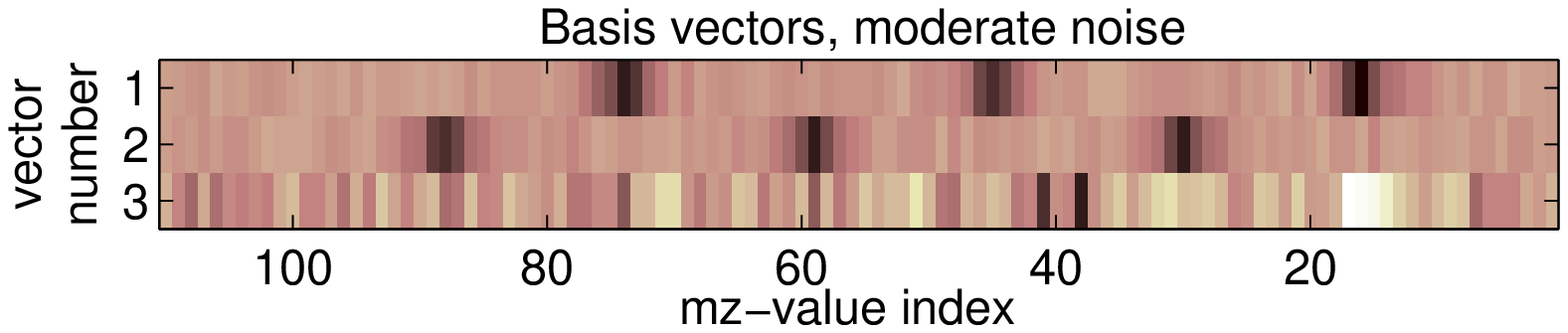}
\end{minipage}}
\hspace{2pt}

\centerline{\begin{minipage}{7pt}
\large \textbf{c} \vspace{80pt}
\end{minipage}  
\begin{minipage}{0.7\textwidth}
\includegraphics[width=\textwidth]{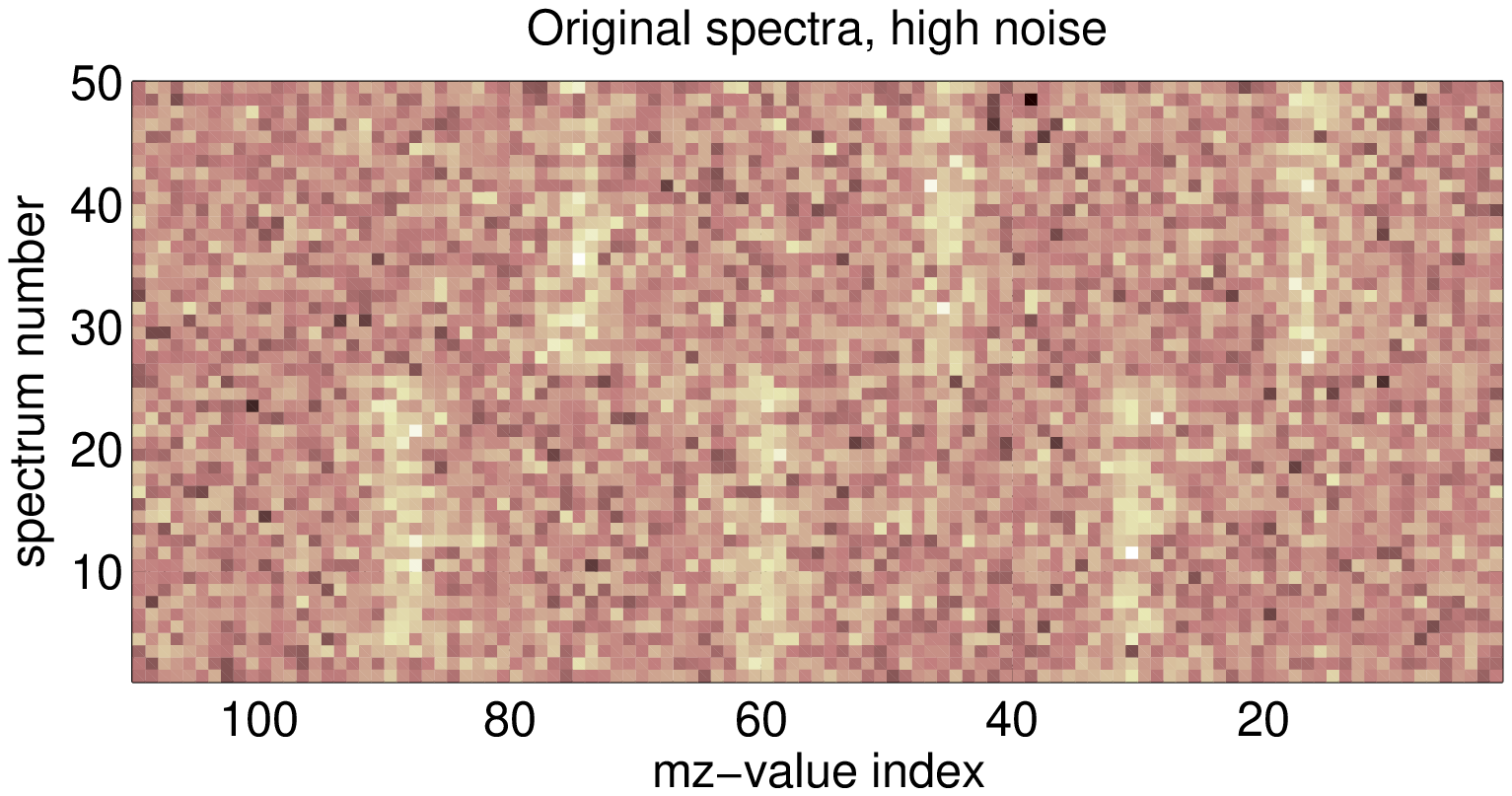}
\end{minipage}}
\hspace{0pt}

\centerline{\begin{minipage}{7pt}
\large \textbf{d} \vspace{30pt}
\end{minipage} 
\begin{minipage}{0.7\textwidth}
\includegraphics[width=\textwidth]{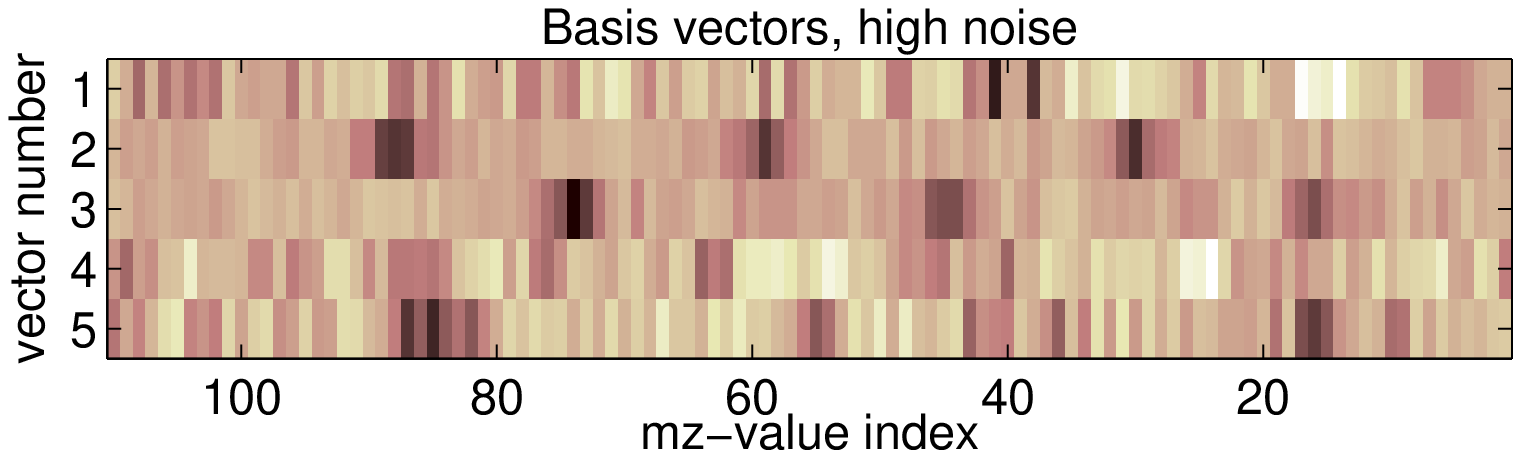}
\end{minipage}}
\caption{Examples of two generated spectra sets (two classes of spectra each with three peaks), 
with moderate noise (\textbf{\normalsize a}) and with high noise (\textbf{\normalsize c})  as well as 
the learned basis vectors (plotted along $mz$-value index), for moderate (\textbf{\normalsize b}) and for high noise settings (\textbf{\normalsize d}).}%
\label{fig:AKSS09_spectramatr}%
\end{figure}

We considered two set of settings corresponding to a high and moderate noise 
which imitate  measurements of a low-resolutional MS instrument. The length of spectra is $L=110$,
which corresponds to the length of only  a part of a real-life spectrum; the number of 
spectra is $R=50$; the number of classes considered is $D=2$; for each class
we pre-specified positions of three characterizing peaks. 
Examples of the simulated spectra sets (moderate and high noise) are 
depicted in Fig.~\ref{fig:AKSS09_spectramatr}. 

The proposed peak picking procedure has three parameters besides the minimal possible peak width, namely 
the $l_1$-regularization parameter $\alpha$,  $l_2$-regularization parameter $\beta$ and 
the constraint on the norm of a basis vector $C$. In this paper we investigate the accuracy 
of the procedure and its dependence on the parameters $\alpha$ and $C$. The role of the parameter
$\beta$ and the strategy for its choice is investigated in~\cite{AKSS09_schifflerSPARS09}.
Although we do not provide rules for the precise choice of the parameters $\alpha$ and $C$, 
in Section~\ref{subsec:simdata} we recommendations on their selection.
Moreover, note that even without a strategy for the choice of parameters one can apply the procedure,
for example when the peak picking is combined with a classification of samples with known labels.
Then the peak picking parameters can be optimized using e.g. cross-validation minimizing 
the classification accuracy.

\subsection{Evaluation results for the simulated data}
\label{subsec:simdata}

\begin{figure}%
\hspace{-0pt}
\centerline{\includegraphics[width=0.99\textwidth]{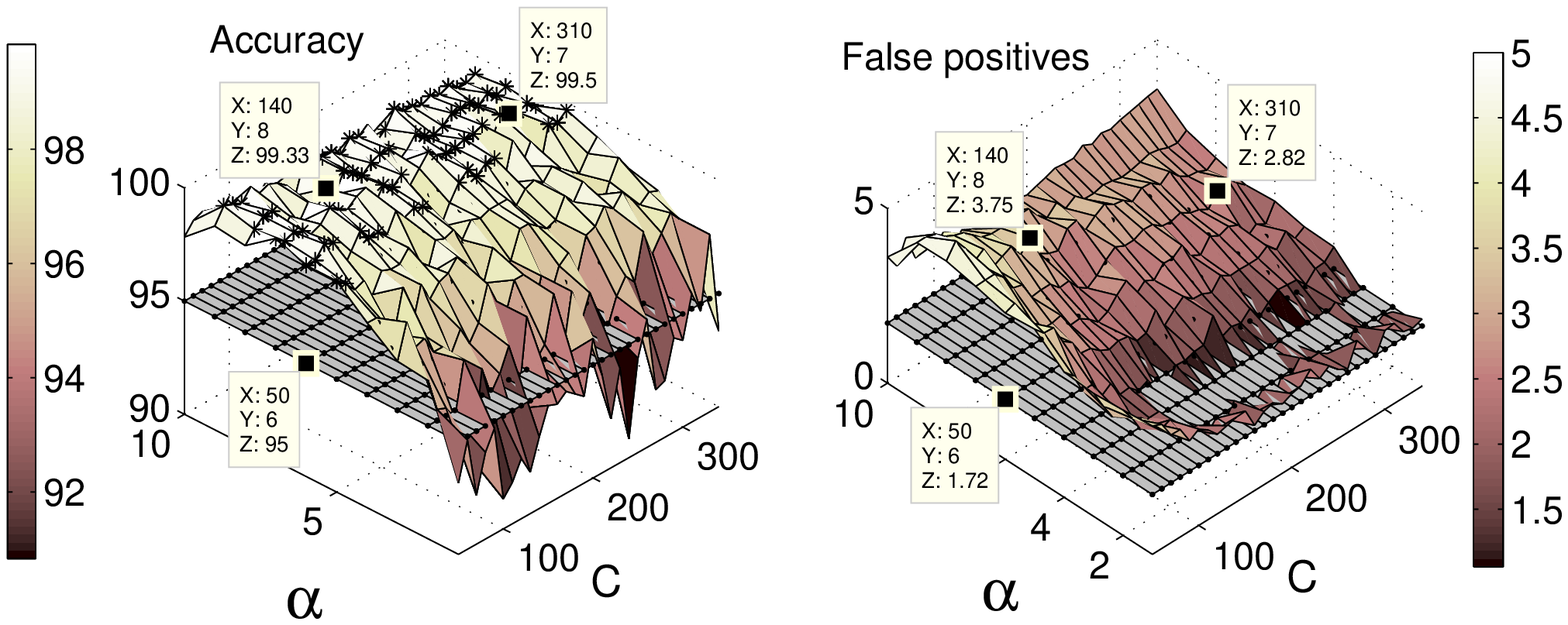}}%
\caption{The mean values of the accuracy of determining the peak positions (left) and 
the number of false positives (right), moderate noise, $\beta=10^{-10}$. 
For better illustration the values at two arbitrary points are given.
The grey horizontal planes correspond to the reference results of the mean spectrum-based procedure. 
The values of accuracy higher than 99\% are marked with stars.}%
\label{fig:AKSS09_GSresults_onlySC}%
\end{figure}

The investigation of the role of the parameters has been done using an extensive grid search 
through 10 values of $\alpha$ (1:1:10) (this notation denotes a grid of values from one to 10 with a step one) and 26 values of $C$ (50:10:300) for four different $\beta$ (1,$10^{-1}$,$10^{-5}$,$10^{-10}$). 
Before the grid search we simulated 100 replicates of spectra sets for high and moderate noise 
(the noise and the positions of spurious peaks are simulated).
Then we calculated the mean values of (1) the accuracy of determining the exact positions of the peaks for each spectrum
(i.e. the ratio between the number of correctly found peaks and six, which is the number of all ground-truth peaks
in a dataset) and (2) the number of false positives (i.e. the number of peaks found which 
are found by the procedure but are not the ground-truth peaks). Fig.~\ref{fig:AKSS09_GSresults_onlySC} shows 
the mean values of the accuracy and the number of false positives for one spectrum for $\beta=10^{-10}$
for the moderate noise settings.

One can see that, as expected, the accuracy mostly depends not on $C$ but on the regularization 
parameter $\alpha$. The same, but to a lesser extent, is true for the number of false positives.
Moreover, the procedure provides almost perfect accuracy (more than 99\%) in a quite large
region of parameters. However, in the same region the number of false positives increases.

In oder to produce reference values for our results we considered another peak picking procedure. 
We calculated the mean spectrum for the given dataset $\mathbf{X}$ and then applied the same
simple peak picking method from section~\ref{sec:AKSS09_peakdetection} to the mean spectrum.
This allows us to compare the feasibility of the sparse coding basis vectors to be exploited
for the peak picking with the standardly used mean spectrum. For the same replicates of 
the dataset with a moderate noise, this procedure provided the mean accuracy of $95\%$ and 
the number of false positives per spectrum equal to $1.72$. These reference results 
are plotted in Fig.~\ref{fig:AKSS09_GSresults_onlySC} as grey horizontal planes.

Comparing these reference results with our results, one can see that the sparse coding-based 
procedure provides better accuracy in a large region of parameters (98\%--99\% vs. 95\%) 
but results in a larger number of false positives (around 3 vs. 1.72). Therefore, we conclude
that the sparse coding procedure leads to more accurate peak picking 
than the mean spectrum-based procedure though is less specific. 
We guess that this is due to the fact that we have two classes and a peak in the mean spectrum
has only a half of its real intensity. 
This was confirmed by experiments with
one class-dataset where the difference between our procedure and the mean spectrum-based
procedure was less than in the two-classes dataset (results are not shown in this paper).

\begin{figure}%
\hspace{-0pt}
\centerline{\includegraphics[width=0.99\textwidth]{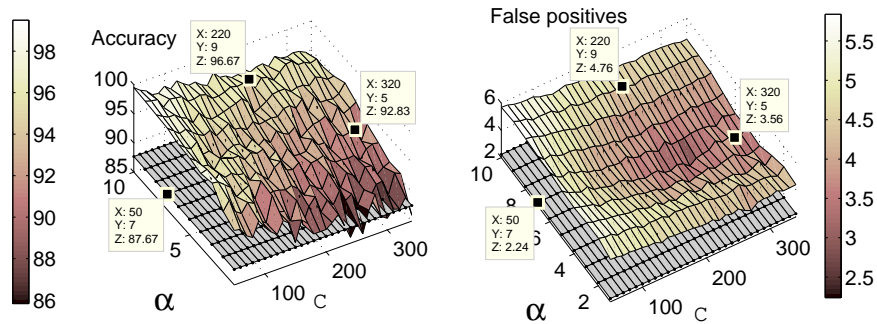}}%
\caption{The mean values of the accuracy of determining the peak positions (left) and the number of false positives (right), 
high noise, $\beta=10^{-10}$.
For better illustration the values at two arbitrary points are given.
The grey horizontal planes correspond to the reference results of the mean spectrum-based procedure.}%
\label{fig:AKSS09_GSresults_onlySC_highnoise}%
\end{figure}

Let us consider the results for the set of spectra with a high noise, see Fig.~\ref{fig:AKSS09_GSresults_onlySC_highnoise}.
Again, as for a moderate noise, our procedure is more accurate than the reference procedure though is less specific.
Moreover, the difference between the reference procedure and our procedure is increased. 
The accuracy of the mean spectrum-based procedure is equal to 88\% (vs. 95\% for a moderate noise) whereas
is 93\%--98\% (vs. 98\%--99\% for a moderate noise)
for our procedure in the region of high values. We conclude that our procedure
is not only more accurate than the mean spectrum-based procedure but is also
more robust to a high noise. 

Based on these results, we suggest the following accuracy-maximizing strategy for the choice of parameters. 
The $l_1$-regularization parameter $\alpha$ should be selected as large as possible (which leads to a high accuracy) 
while the procedure returns reasonable amount of basis vectors. For instance, under assumption of two classes, 
one should increase $\alpha$ as long as the procedure returns not less than two vectors.
The choice of $C$ is not crucial. The $l_2$-regularization parameter $\beta$ should
be as small as possible as proposed in~\cite{AKSS09_schifflerSPARS09}.

\subsection{Real-life dataset: colorectal cancer}
\begin{figure}%
\hspace{0pt}
\hspace{0pt}

\centerline{\begin{minipage}{7pt}
\large \textbf{a} \vspace{60pt}
\end{minipage} 
\begin{minipage}{0.67\textwidth}
\includegraphics[width=\textwidth]{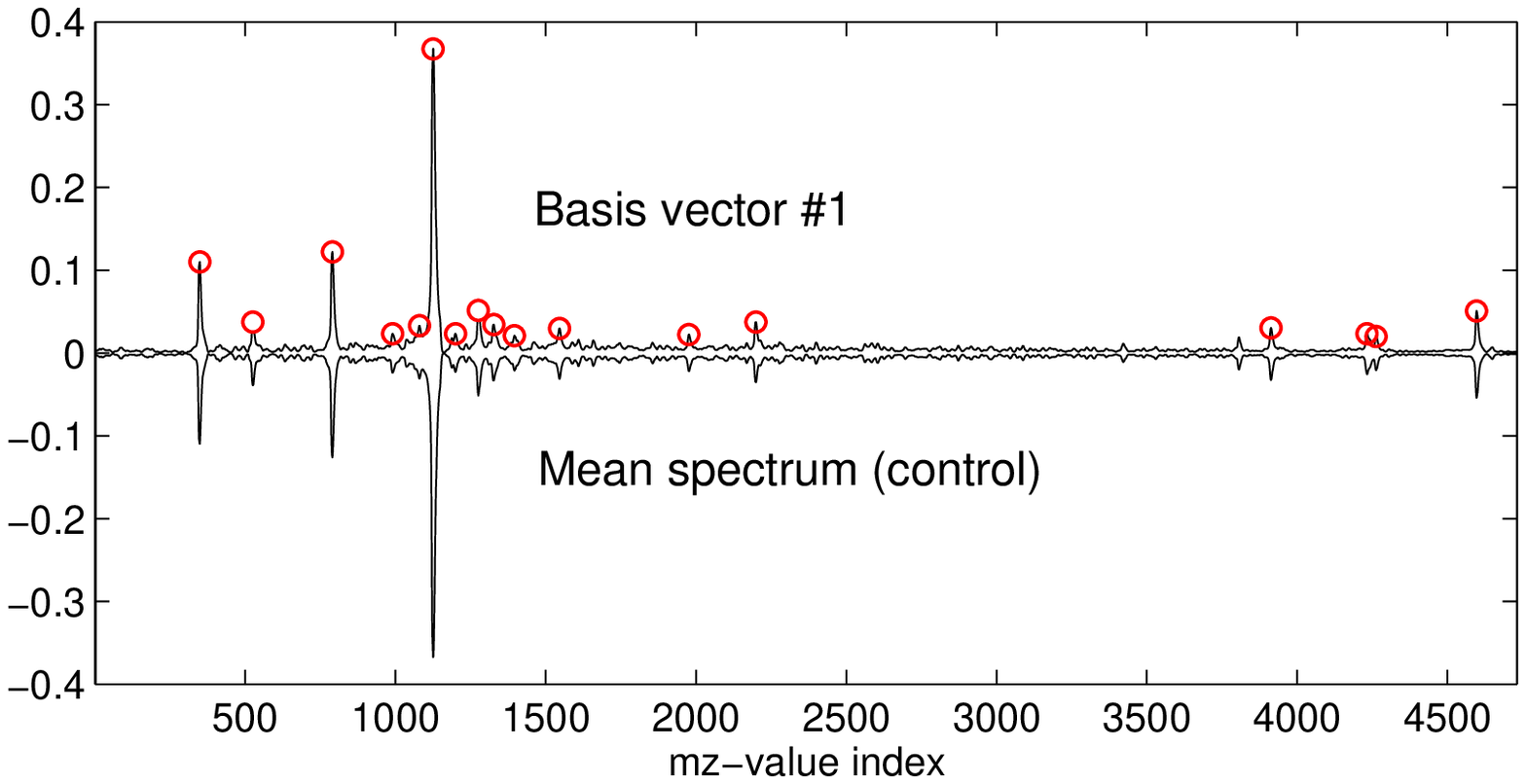}
\end{minipage}}
\hspace{0pt}

\centerline{\begin{minipage}{7pt}
\large \textbf{b} \vspace{60pt}
\end{minipage} 
\begin{minipage}{0.67\textwidth}
\includegraphics[width=\textwidth]{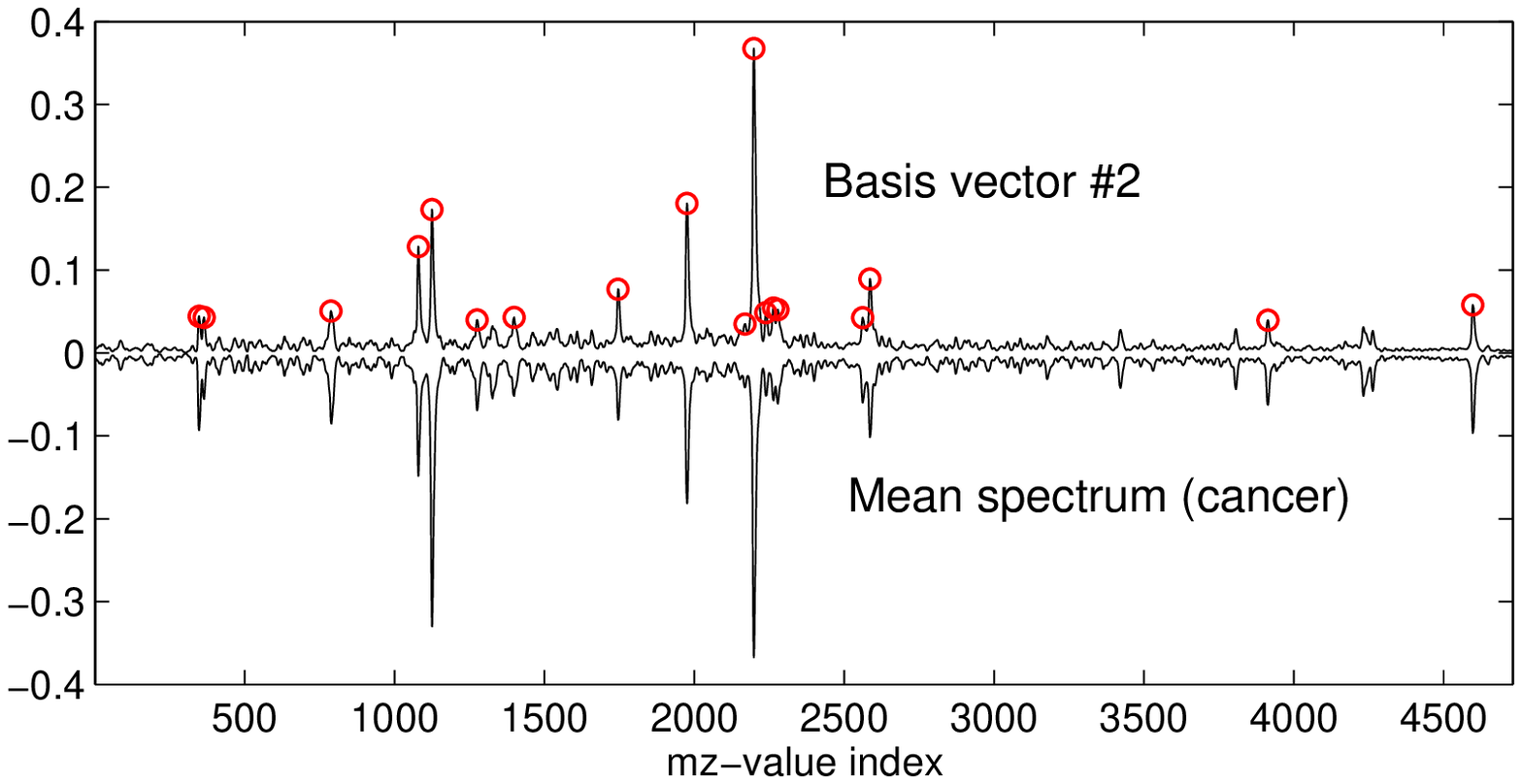}
\end{minipage}}
\caption{Colorectal cancer results. First basis vector and the mean control spectrum (\textbf{\normalsize a}). 
Second basis vector and the mean cancer spectrum (\textbf{\normalsize c}). Found peaks are marked with circles. 
Mean spectra are attributed to corresponding basis vectors manually, then scaled and plotted with negative sign.}%
\label{fig:AKSS09_realdata1}%
\end{figure}

We tested the proposed peak picking procedure on a real-life dataset~\cite{AKSS09_noo06} of 64 colorectal 
cancer MALDI-TOF (matrix-assisted laser desorption/ionization) spectra and 48 control spectra considered 
in the region [1100,3000]~Da containing 4731 $mz$-values (spectrum length $L=4731$). 
The parameter $\alpha=120$ was selected as proposed above (under assumption of two classes), five different values of $C$ 
(500:500:2500) have been tested producing similar results, $\beta=10^{-10}$. The 
basis vectors for $C=500$ are shown in Fig.~\ref{fig:AKSS09_realdata1}. One can see that 
the basis vectors resulted after the unsupervised learning can precisely catch the patterns of the per-class mean spectra.
\texttt{}

\subsection{Real-life dataset: liver diseases}
\begin{figure}%
\hspace{0pt}
\hspace{0pt}

\centerline{\begin{minipage}{7pt}
\large \textbf{a} \vspace{60pt}
\end{minipage} 
\begin{minipage}{0.67\textwidth}
\includegraphics[width=\textwidth]{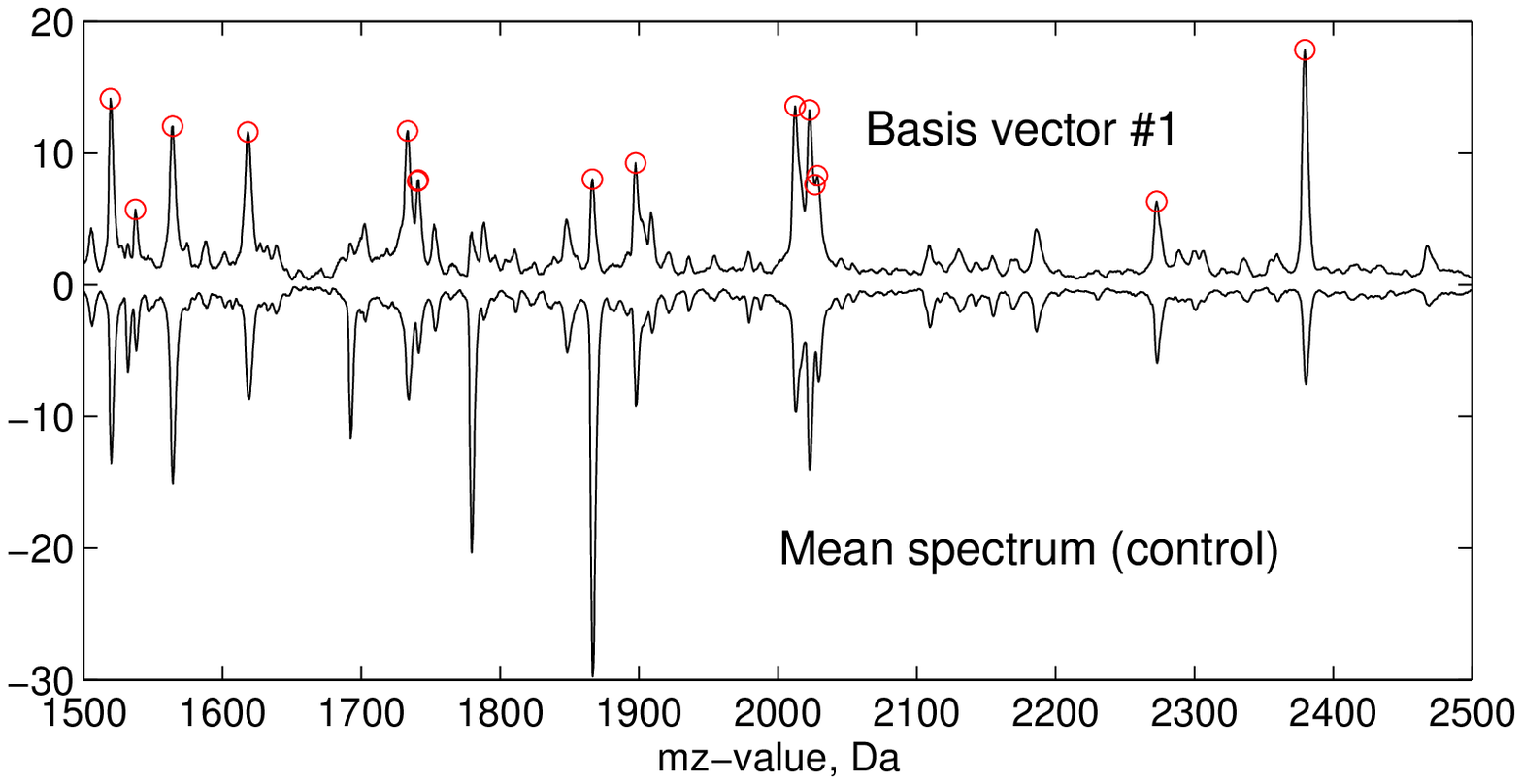}
\end{minipage}}
\hspace{0pt}

\centerline{\begin{minipage}{7pt}
\large \textbf{b} \vspace{60pt}
\end{minipage} 
\begin{minipage}{0.67\textwidth}
\includegraphics[width=\textwidth]{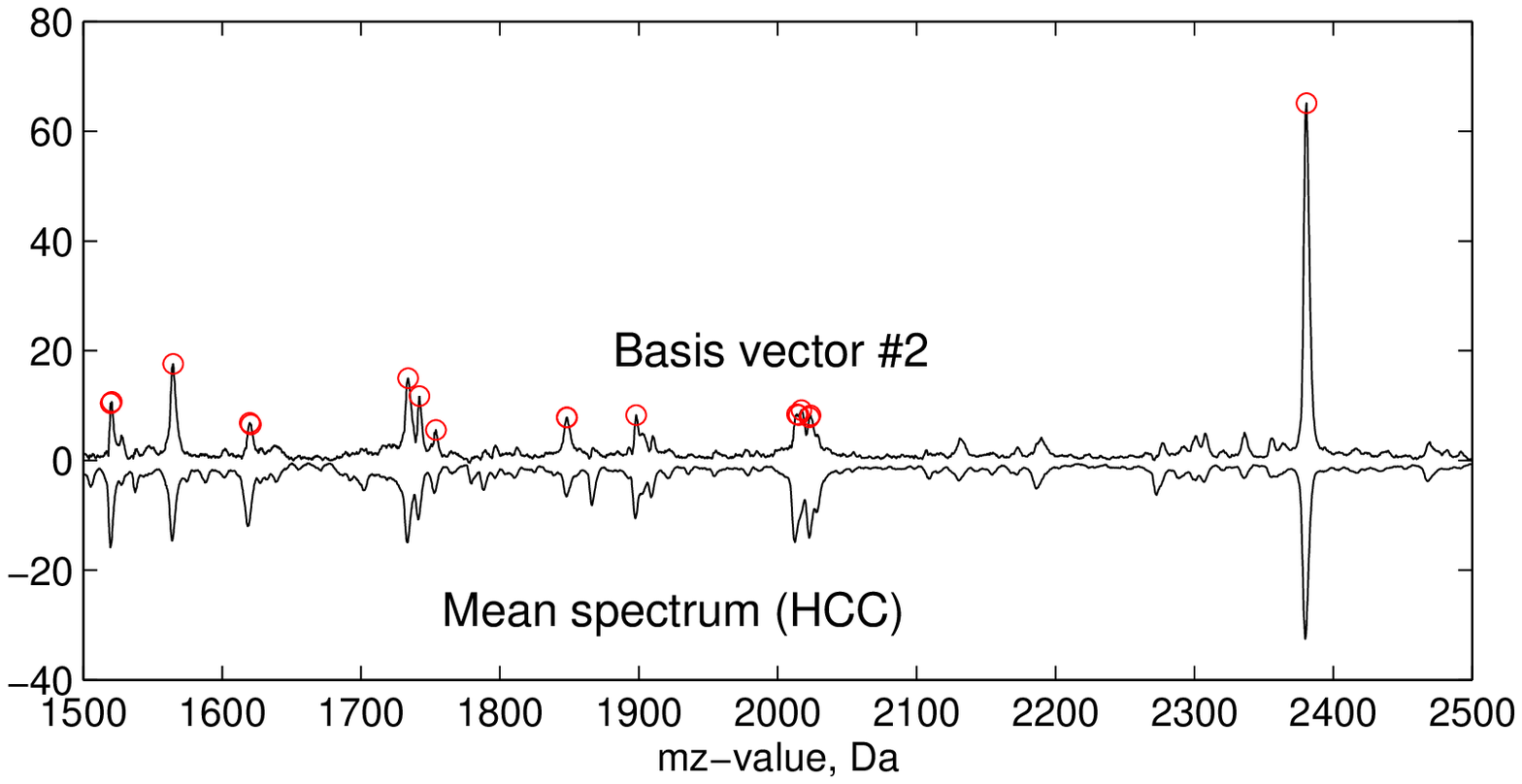}
\end{minipage}}
\hspace{0pt}

\centerline{\begin{minipage}{7pt}
\large \textbf{c} \vspace{60pt}
\end{minipage} 
\begin{minipage}{0.67\textwidth}
\includegraphics[width=\textwidth]{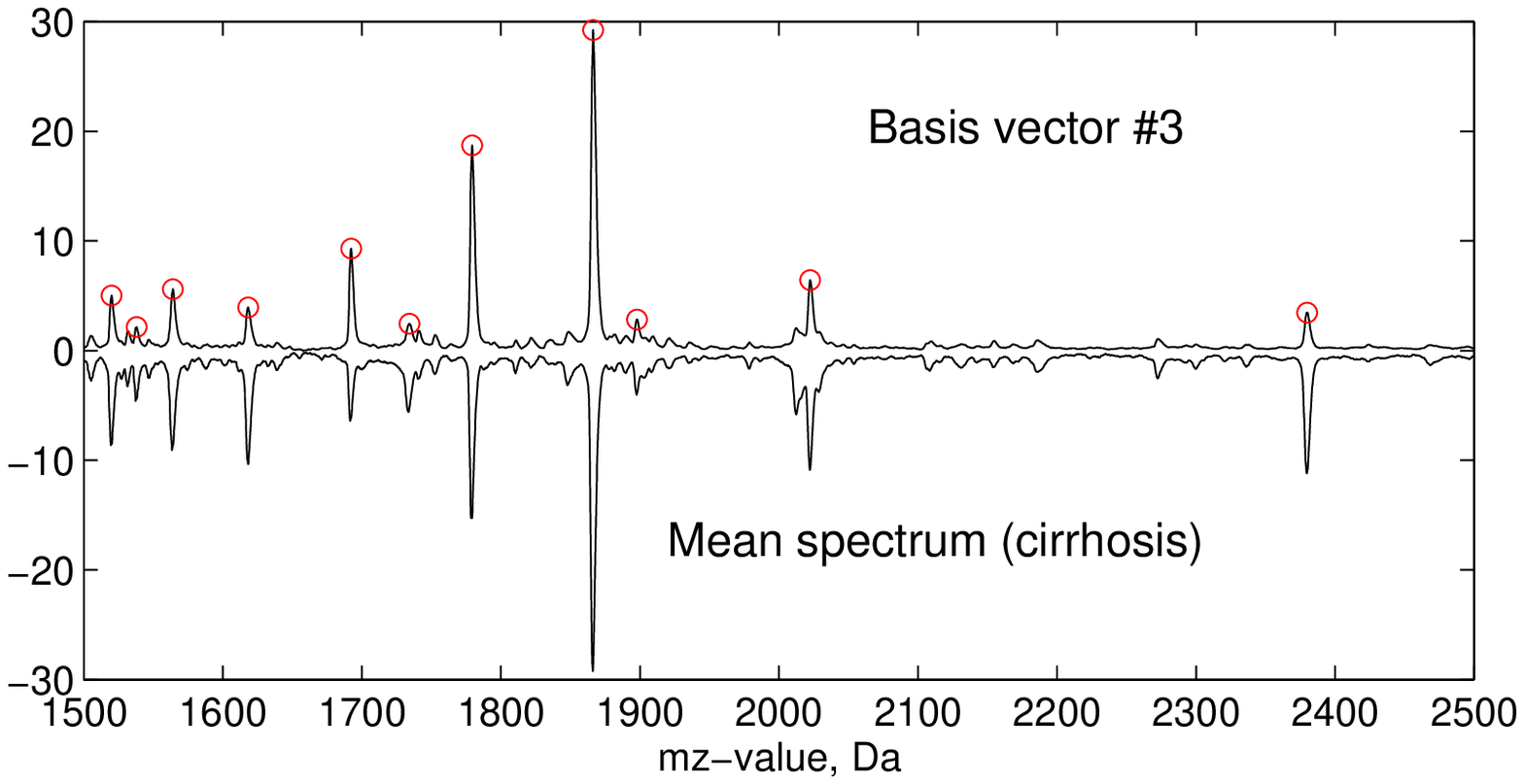}
\end{minipage}}
\caption{Liver diseases results. First (\textbf{\normalsize a}), second (\textbf{\normalsize b}) and third (\textbf{\normalsize c}) basis vectors plotted with mean spectra of control, HCC and cirrhosis classes. Found peaks are marked with circles. 
Mean spectra are attributed to the basis vectors manually, then scaled and plotted with negative sign.}%
\label{fig:AKSS09_realdata2}%
\end{figure}

In order to illustrate application of our procedure to a set of spectra of more than two classes, 
we examined a liver diseases dataset of \cite{AKSS09_Ressom07} with mass spectra corresponding to 
78 hepatocellular carcinoma (HCC), 51 cirrhosis, and 72 control serum samples. For our purpose we considered only the region [1500,2500]~Da
with 5108 $mz$-values (spectrum length $L=5108$). This region was selected since the mean spectra
of the three classes in this region are quite different (see Fig.~\ref{fig:AKSS09_realdata2}). 
In \cite{AKSS09_Ressom07} significant peaks were found in the region [7500,8500]~Da. However,
in \cite{AKSS09_Ressom07} another two-classes classification problem of comparison disease spectra
(HCC and cirrhosis jointly) with control spectra was considered. The parameter $\alpha=94.5$ was selected 
as proposed above (under assumption of three classes), five different values of $C$ 
(500:500:2500) have been tested producing similar results, $\beta=10^{-10}$. Again, as 
for two-classes real-life dataset, the basis vectors are very similar to the per-class mean spectra

\section{Conclusions}
\label{sec:AKSS09_concl}
The main contribution of this paper is a way of sparse representation of mass spectra,
which we propose to use for the peak picking. We have found
that the obtained representation basis is similar to per-class mean spectra.
In contrast to the mean spectra, this basis can be obtained
in an unsupervised manner. 
In this paper a basic peak picking method applied
to the basis vectors was used to demonstrate the potential of sparse coding. 
We expect that application of a more advanced peak picking method
will lead to even better results.


\end{document}